\documentclass[10pt,journal,compsoc,twocolumn]{IEEEtran}
\ifCLASSOPTIONcompsoc
  \usepackage[nocompress]{cite}
\else
  \usepackage{cite}
\fi

\ifCLASSINFOpdf
   \usepackage[pdftex]{graphicx}
   \graphicspath{{../pdf/}{../jpeg/}}
   \DeclareGraphicsExtensions{.pdf,.jpeg,.png}
\else
   \usepackage[dvips]{graphicx}
   \graphicspath{{../eps/}}
   \DeclareGraphicsExtensions{.eps}
\fi

\ifCLASSOPTIONcompsoc
 \usepackage{subfigure}
\else
 \usepackage{subfigure}
\fi

\usepackage{subfigure}
\usepackage{CJKutf8}
\usepackage{extarrows}
\usepackage{multirow}
\usepackage{hyperref}
\usepackage{url}
\usepackage{booktabs}

\usepackage{url}
\usepackage{stfloats}
\usepackage{booktabs}
\usepackage{multirow}
\usepackage[ruled]{algorithm2e}
\usepackage{algpseudocode}
\usepackage{array}
\usepackage{color}
\usepackage{amsmath}
\usepackage{amssymb}
\usepackage{bm}

\usepackage{subfigure}

\begin{document}
\title{Uncertainty-Aware Relational Graph Neural Network for Few-Shot Knowledge Graph Completion}



\author{Qian~Li,
        Shu Guo,
        Yinjia Chen,
        Cheng Ji,
        Jiawei Sheng,
        and
        Jianxin Li
        \IEEEmembership{}
\IEEEcompsocitemizethanks{\IEEEcompsocthanksitem Q. Li and J. Li
are with Beijing Advanced Innovation Center for Big Data and Brain Computing, Beihang University, Beijing 100083, and also with the State Key Laboratory of Software Development Environment, Beihang University, Beijing 100083, China. 
E-mail: \{liqian, lijx,  jicheng\}@act.buaa.edu.cn
\IEEEcompsocthanksitem S. Guo is the National Computer Network Emergency Response Technical Team/Coordination Center of China, Beijing 100083, China. 
E-mail: guoshu@cert.org.cn.
\IEEEcompsocthanksitem Y. Chen is with the School of Mathematics, Edinburgh University, Edinburgh, UK. E-mail: chenyinjia2000@gmail.com.
\IEEEcompsocthanksitem J. Sheng is with the Institute of Information Engineering, Chinese Academy of Sciences, Beijing 100083, China, and the School of Cyber Security, University of Chinese Academy of Sciences, Beijing 100083, China. E-mail: shengjiawei@iie.ac.cn.
}
\thanks{Manuscript received April 2023. (Corresponding author: Jianxin Li.)}
}


\markboth{IEEE Transactions on Knowledge and Data Engineering, ~Vol.~14, No.~9, 
November~2023}%
{Shell \MakeLowercase{\textit{et al.}}: Bare Demo of IEEEtran.cls for Computer Society Journals}

\IEEEtitleabstractindextext{%
\begin{abstract}
Few-shot knowledge graph completion (FKGC) aims to query the unseen facts of a relation given its few-shot reference entity pairs. 
The side effect of noises due to the uncertainty of entities and triples may limit the few-shot learning, but existing FKGC works neglect such uncertainty, which leads them more susceptible to limited reference samples with noises.
In this paper, we propose a novel uncertainty-aware few-shot KG completion framework (UFKGC) to model uncertainty for a better understanding of the limited data by learning representations under Gaussian distribution.
Uncertainty representation is first designed for estimating the uncertainty scope of the entity pairs after transferring feature representations into a Gaussian distribution.
Further, to better integrate the neighbors with uncertainty characteristics for entity features, we design an uncertainty-aware relational graph neural network (UR-GNN) to conduct convolution operations between the Gaussian distributions. Then, multiple random samplings are conducted for reference triples within the Gaussian distribution to generate smooth reference representations during the optimization. 
The final completion score for each query instance is measured by the designed uncertainty optimization to make our approach more robust to the noises in few-shot scenarios. 
Experimental results show that our approach achieves excellent performance on two benchmark datasets compared to its competitors.

\end{abstract}

}

\maketitle

\IEEEpeerreviewmaketitle


\IEEEPARstart{K}nowledge graphs (KGs) represent information pieces consisting of entities and their relations, usually organized in the form of triples (head entity, relation, and tail entity). 
KGs have been proven effective for multiple downstream NLP tasks such as dialogue systems \cite{DBLP:conf/sigir/KimL22, DBLP:conf/sigir/Sun0BRRCR21, DBLP:conf/acl/HeBEL17}, entity search \cite{DBLP:conf/sigir/GerritseHV22, DBLP:journals/kais/Komamizu20, DBLP:conf/sigir/ShenXHSS018}, and recommendation~\cite{DBLP:conf/sigir/0002ZCZG22, DBLP:conf/sigir/WangZ0ZW022}.
Although KGs usually contain abundant triples, they still suffer from incompleteness problems that many entities and relations are lost.
The KG completion task aims to infer the missing facts (e.g., tail entities) by analyzing the available triple sets in the same KG but requires sufficient training triples to represent each relational learning expression \cite{DBLP:conf/sigir/ChenZLDTXHSC22, DBLP:conf/coling/SongHZG00022, DBLP:conf/emnlp/ZhangLJLWJY21}.
In the real-world scenarios, relations in the already-existing KGs generally suffer from the long-tail problem \cite{DBLP:conf/sigir/NiuLTGDLWSHS21, DBLP:conf/emnlp/WangLLBL19, xiong-etal-2018-one}, meaning that a large number of relations only occur occasionally with insufficient training samples in a KG, which makes the traditional KG completion methods impractical. Therefore, the few-shot KG completion (FKGC) task, which predicts the tail entity from a given head entity and a specific relation as well as a few triples for the relation, has gained wide attention.

\begin{figure}[t]
    \centering
    \includegraphics[width=\linewidth]{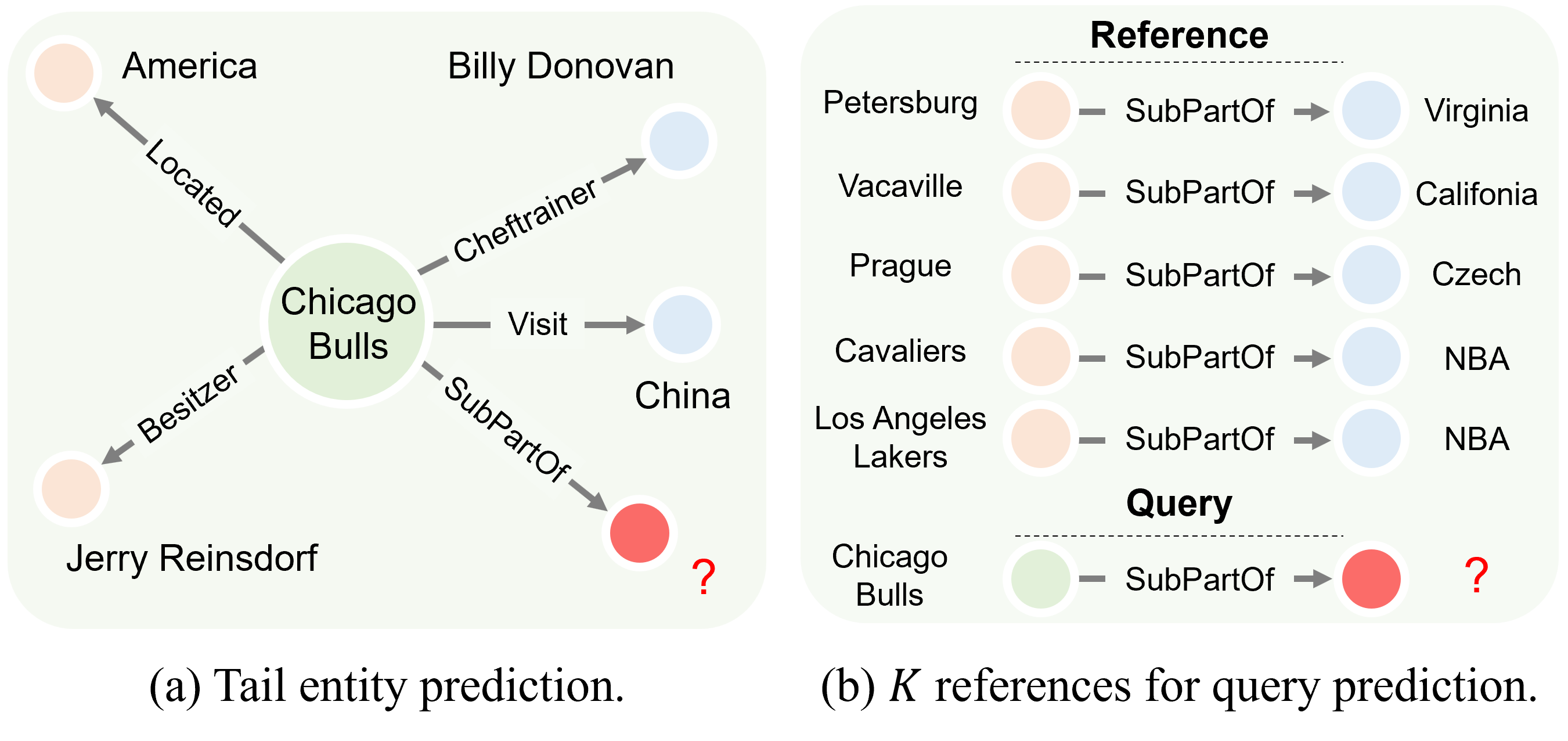}
    \caption{An example of the few-shot KG completion task. (a) The task is to predict tail entities on KG. (b) The task utilizes only $K$ triples for the specific relation to predicting others.
}
    \label{example}
\end{figure}


Few-shot knowledge graph completion refers to the task of predicting tail entities in a knowledge graph, given only a few examples, as shown in Figure \ref{example}(a). The task involves a reference/support set and a query set, where $K$ known triples are provided as reference for each relation, such as the \textit{SubPartOf} relation shown in Figure \ref{example}(b). The objective is to predict the tail entity for the \textit{SubPartOf} relation, given a head entity \textit{Chicago Bulls} and the relation itself, using the query set. This task is challenging due to the need for the model to generalize effectively to new, unseen triples based on limited training data (i.e., reference triples).

Several latest research works~\cite{xiong-etal-2018-one, DBLP:conf/aaai/ZhangYHJLC20, DBLP:conf/emnlp/ShengGCYWLX20} reveal that good feature embedding is important to deliver favorable performance for the similarity-based classification. They concentrate on developing a one-hop neighbor encoder that utilizes local neighbors and its structure to learn better entity representations.
To further take advantage of the local subgraph structure for entity learning, P-INT~\cite{DBLP:conf/emnlp/XuZKDCLL21} leverages paths between entities to represent complex relations and calculates their interactions.
But they disregard the impact of complex relations. To settle this, MetaR \cite{chen-etal-2019-meta} and GANA~\cite{DBLP:conf/sigir/NiuLTGDLWSHS21} adopt the model agnostic meta-learning (MAML) as the architecture to extract relation-specific information. Later, MetaP \cite{DBLP:conf/sigir/JiangGL21} further addresses this issue by leveraging convolutional networks to capture relation patterns inside an entity pair. 
However, all of these methods represent entities as vectors, i.e., deterministic points in the vector space, which neglects the uncertainties of entities in KGs and further constrains their capability of fully employing existing limited reference triples. Intuitively, an entity containing fewer triplets has more uncertainty. For each few-shot task where each relation links a small number of triplets, the side effect of noises due to the high uncertainty can be more severe in few-shot scenarios. The higher the uncertainty, the less reliable the reference triples. Therefore, modeling uncertainty is especially vital for few-shot KG learning with limited samples, but still under-explored.

To overcome the above challenge, we propose an \underline{U}ncertainty-aware \underline{F}ew-Shot \underline{K}nowledge \underline{G}raph \underline{C}ompletion framework, termed as UFKGC\footnote{The source code is available at \url{https://XX}.}, to model uncertainty for a better understanding of the limited data and expand the data under the same distribution. 
Specifically, for each entity and triple, we convert them from a deterministic scalar to a distribution whose variance characterizes the scope of uncertainty, which can better understanding of entities and triples on limited data.
Uncertainty representation is designed to evaluate the uncertainty of pairs and has different uncertainties of different pairs, making few-shot learning more robust.
Further, to better integrate the neighborhood information for entity features, we design an uncertainty-aware relational graph neural network (UR-GNN) to do convolution operations between Gaussian distributions, considering the uncertainty of entities and triples.
To do sample enhancement for reference triples, we conduct multiple random sampling within the Gaussian distribution. Without destroying the original data distribution, we can prompt the model to learn consistent knowledge from those generated smooth representations.
The final completion score for each query instance is measured by the designed uncertainty optimization to be robust on the few-shot setting. 
It promotes few-shot optimization under constrained sample scenarios. Experimental results show that our model achieves SOTA results on the two FKGC benchmark datasets.
Our contributions can be summarized as follows: 
\begin{itemize}
    \item To our best knowledge, we are the first to explore and model the uncertainty of the entities and triples in few-shot knowledge graph completion, where a better understanding of limited data is particularly important. 
    \item We further design the UR-GNN to better integrate the uncertainty information and uncertainty optimization to estimate uncertainty of completion scores.
    \item Experimental results indicate the framework achieves state-of-the-art performance on two public FKGC datasets.
\end{itemize}

\begin{figure*}[t]
    \centering
    \includegraphics[width=\linewidth]{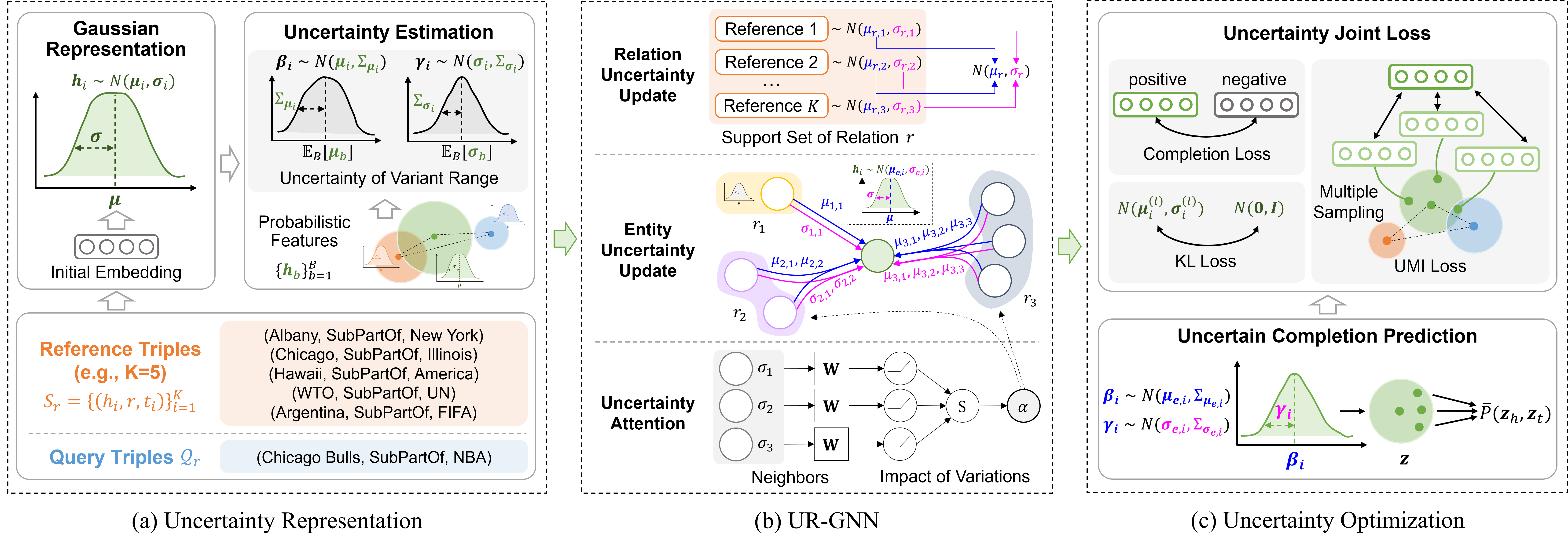}
    \caption{The framework of UFKGC.
    It contains three modules: Uncertainty Representation, UR-GNN, and Uncertainty Optimization. (a) The Uncertainty Representation module is used to convert the deterministic embedding into an uncertain embedding and evaluate the scope of uncertainty. (b) The UR-GNN is to learn embeddings in the uncertain space,
    and (c) Uncertainty Optimization is to evaluate the completion prediction.}
    \label{Framework}
\end{figure*}

\section{Related Work}

\subsection{Few-Shot Knowledge Graph Completion}

Many studies \cite{DBLP:conf/sigir/NiuLTGDLWSHS21, DBLP:journals/corr/abs-2203-11639} make an effort to do few-shot relational modeling regarding long-tail relations to complete KGs. Recent studies can be generally divided into three categories:  
(a) \textit{Metric-based methods}, which attempt to learn a metric by calculating the similarity between support and the query triples. To gain better performance based on this principle, they generally focus on learning informative and robust representations.
The first study on FKGC, known as GMatching \cite{xiong-etal-2018-one}, takes advantage of one-hop neighbors to discover more accurate entity embeddings.
Based on GMatching, FSRL \cite{DBLP:conf/aaai/ZhangYHJLC20} and FAAN \cite{DBLP:conf/emnlp/ShengGCYWLX20} capture local structures with an attention mechanism to aggregate relation-specific neighborhoods from K reference triples. On top of these studies, REFORM \cite{DBLP:conf/cikm/WangHCWL21} further constructs a query-oriented graph, using GCN to assign confidence scores to different triples to mitigate the negative effect brought by errors, and YANA \cite{DBLP:conf/ksem/LiangZCYY22} devises a local pattern graph and adopts GNN to capture the integrate latent relations among the entities of the support and query set. 
P-INT \cite{DBLP:conf/emnlp/XuZKDCLL21} construct directed subgraphs between head and tail entities to represent triples and calculate interactions between support and query sets. On the other hand, MetaP \cite{DBLP:conf/sigir/JiangGL21} passes triples through convolutional layers to extract relation patterns for further comparison.
(b) \textit{Optimization-based methods}, which attempt to learn faster with model agnostic meta-learning (MAML).
The core design of MetaR~\cite{chen-etal-2019-meta} and GANA \cite{DBLP:conf/sigir/NiuLTGDLWSHS21} is built on MAML for accelerating the learning of relation meta and hyper-plane parameters, while Meta-KGR \cite{DBLP:conf/emnlp/LvGHHLL19} integrates MAML with multi-hop pathfinding in learning a better policy for entity selection. 
(c) \textit{Cognitive graph-based method}, specifically CogKR~\cite{du2021cogkr}, which finishes the completion by introducing the dual process from cognitive science to construct and update a cognitive graph. 
Nevertheless, the above existing methods ignore modeling the inherent uncertainty of features for more effective representations. 

\subsection{Uncertainty on Deep Learning}

The uncertainty methods in deep learning transfer a fixed-length feature vector into a Gaussian distribution to model uncertainty.
Variational Auto-encoder~\cite{DBLP:journals/corr/KingmaW13}, an important method for modeling data uncertainty, has been getting more attention in deep learning. 
\cite{DBLP:journals/corr/VilnisM14} represents words using Gaussian embeddings, where the covariance of the words naturally indicates their ambiguity. 
In addition, TIM~\cite{DBLP:conf/nips/BoudiafZRDPA20} measures the uncertainty of the model based on the uncertainty of similarity through mutual information. They indicate that the Gaussian embedding method is an effective way to model uncertainty.
To learn the uncertainty features of KGs, \cite{DBLP:conf/cikm/HeLJ015} attempts to represent the entities and relations in KGs with the Gaussian distribution, while \cite{DBLP:conf/dasfaa/ZhangWQ21} and \cite{wang2022incorporating} continue to explore this idea in the few-shot uncertain KG completion(FUKGC) task. It is noteworthy that the FUKGC task introduces an additional measure of triple confidence, which \cite{DBLP:conf/dasfaa/ZhangWQ21} and \cite{wang2022incorporating} assume as the optimization objective of triple covariance. In contrast, the few-shot knowledge graph completion task lacks access to such information.
From another perspective, \cite{wang2019tackling} proposes to generate new samples for relieving the data sparsity problem by modeling the sample triples in Gaussian distribution with Variational Auto-encoder.
Together, these works indicate that representing the uncertainty of KGs might boost the effectiveness and robustness of the learning algorithms.

In this paper, we model the uncertainty of knowledge information for the few-shot knowledge graph completion, promoting the exploitation of limited samples in few-shot optimization. To the best of our knowledge, this is the first study attempting to apply uncertainty modeling to few-shot knowledge graph completion.

\section{Preliminaries}
We first provide the definitions of the knowledge graph and the few-shot knowledge graph completion (FKGC) as follows.

\textbf{Knowledge Graph.}
A KG $\mathcal{G}=\{\mathcal{E}, \mathcal{R}, \mathcal{T}\}$ is a group of factual information, where $\mathcal{E}$ and $\mathcal{R}$ represent the sets of entities and relations, respectively. All the triples in $\mathcal{G}$ are included in the triple set $\mathcal{T}=\{(h, r, t) \in \mathcal{E} \times \mathcal{R} \times \mathcal{E}\}$, where $h$ and $r$ denote the head and tail entity respectively, and $r$ denotes the relation.

\textbf{Few-Shot KG Completion (FKGC).}
The FKGC task aims at predicting possible new triples, given the arbitrary two elements in a triple with just a few training triples available.
We focus on entity completion, specifically forecasting the potential tail entity $t$ from a given head entity $h$ and a few-shot relation $r \in \mathcal{R}$.

In FKGC, one $K$-shot relation corresponds to one task. Given a $K$-shot relation $r\in \mathcal{R}$, $\mathcal{D}_r=\left\{\mathcal{S}_r,\mathcal{Q}_r\right\}$ denotes the $K$-shot completion task w.r.t. relation $r$, consisting of a support set $\mathcal{S}_r$ and a query set $\mathcal{Q}_r$. The support set $\mathcal{S}_r=\left\{(h_i,r,t_i)\in \mathcal{T}\right\}_{i=1}^{K}$ is composed of $K$ samples, and the query set $\mathcal{Q}_r=\left\{(h_j,r,t_j)\in \mathcal{T}\right\}_{j=1}^{Q}$ is composed of several entity pairs for prediction, with $h_j$ as the query head entity and $t_j$ as the ground truth tail entity for the query. 
The relations in $\mathcal{R}$ are divided into three parts, $\mathcal{R}_{train}$, $\mathcal{R}_{val}$, and $\mathcal{R}_{test}$, for the training set, validation set, and testing set, respectively.

\section{Framework}

We present a novel approach called Uncertainty-Aware Few-Shot Knowledge Graph Completion (UFKGC), which is designed to address the challenge of modeling uncertainty in the context of knowledge graph completion. UFKGC leverages the power of Gaussian distribution to capture and represent the uncertainty associated with both entities and relations.
As shown in Figure \ref{Framework}, our model comprises three main parts: 

(1) \textbf{Uncertainty Representation.} The uncertainty of the entities/relations is estimated using the uncertainty representation in the form of Gaussian distribution. Furthermore, it exploits the variance of the statistics of the features to offer guidelines in multi-variate Gaussian distribution. 

(2)~\textbf{Uncertainty Relational Graph Neural Network (UR-GNN).} To better integrate the uncertainty neighbors and relations for entity features, we further design the UR-GNN making few-shot learning more robust. It executes convolution operations between Gaussian distributions and implements variance-based attention to give neighbors varied weights. 

(3) \textbf{Uncertainty Optimization.} To alleviate the adverse effects of increased uncertainty caused by limited samples, we propose to use multiple sampling from the learned representation distribution to measure the completion performance. To be robust to the noises in few-shot settings, we further propose the uncertainty joint loss, which includes completion loss, uncertainty mutual information loss, and KL-divergence loss to jointly evaluate the performance.

In general, our model is proposed for modeling the uncertainty of entities and relations exacerbated by the limited number of samples.
It transfers the feature representations into a Gaussian distribution with uncertainty estimation. The UR-GNN is next designed to update the distributions of entities and relations. To evaluate the performance of the uncertainty representation, we formulate the additional uncertainty mutual information loss and KL-divergence loss for prediction evaluation.

\subsection{Uncertainty Representation}


In order to effectively capture and model uncertainty for the task of few-shot Knowledge Graph (KG) completion, we design a method for representing uncertainty within our framework. This approach involves the utilization of a multi-variate Gaussian distribution to represent the features of entities and relations.
It enhances the precision and expressiveness of our uncertainty representation, ultimately leading to more accurate and reliable predictions in the context of knowledge graph completion tasks, even when dealing with limited data or novel entities and relations.

\subsubsection{Gaussian Representation}
To preserve semantic information, the entities and relations are initially initialized using BERT \cite{DBLP:conf/nips/VaswaniSPUJGKP17}. However, it is impractical to rely on deterministic representations learned from a limited number of few-shot samples. Therefore, we adopt a probabilistic approach to model uncertainty.
Specifically, we choose to represent each feature (whether it belongs to an entity or a relation) using a Gaussian distribution. Let $\boldsymbol{h}_i \sim N(\boldsymbol{\mu}_i, \boldsymbol{\sigma}_i)$ represent the $i$-th feature, where $i$ ranges from 1 to $N_e+N_r$. Here, $N_e$ and $N_r$ denote the total number of entities and relations, respectively.
To fully characterize the Gaussian distribution, we need to determine the mean vector $\boldsymbol{\mu}_i$ and the variance matrix $\boldsymbol{\sigma}_i^2$ associated with each feature. The mean vector $\boldsymbol{\mu}_i$ represents the central value or the expected feature value for the $i$-th feature. On the other hand, the variance matrix $\boldsymbol{\sigma}_i^2$ captures the variability or uncertainty of the feature's potential values.
\begin{equation}
\boldsymbol{\mu}_i=\mathbf{W}_\mu \boldsymbol{f}_i, 
\end{equation}
\begin{equation}
\boldsymbol{\sigma}_i=\mathbf{W}_\sigma \boldsymbol{f}_i,
\end{equation}
where $\boldsymbol{f}_{i}$ refers to the initial representation of the entity or relation (i.e., $\boldsymbol{f}_{i}^{e}$ or $\boldsymbol{f}_{i}^{r}$). $\mathbf{W}_\mu$ and $\mathbf{W}_\sigma$ are variables that can be trained. 
The former $\boldsymbol{\mu}_i$ behaves as the regular feature vector would in a traditional model, while the latter $\boldsymbol{\sigma}_i$ quantifies the uncertainty in features. The larger the variance of the association, the more uncertain the features. By incorporating such probability distributions for each feature, we can model the uncertainty inherent in the representation. This approach allows us to capture a range of possible feature values, rather than relying on a single deterministic representation. By considering the inherent uncertainty, we can account for potential variations and make more informed decisions based on the probability distribution. By incorporating probability distributions, we can better address uncertainty in the feature space and make more robust and reliable decisions based on the inherent variability.

\subsubsection{Uncertainty Estimation}

To accurately portray the uncertainty scope of each probabilistic feature statistic, it is essential to perform uncertainty calculations that consider the uncertainty of the variant range. In light of this, we propose the implementation of an uncertainty estimation module that takes into account potential uncertainties. Our approach involves utilizing a multi-variate Gaussian distribution to represent the features of entities and relations.
In our proposed module, the feature statistics' mean and standard deviation have a probabilistic representation derived from a specific distribution. The center of the Gaussian distribution represents the original statistics of each feature, while the standard deviation represents the uncertainty range for possible shifts. By incorporating this probabilistic technique, we can train models to become more resilient to statistical changes. This is achieved by randomly selecting a variety of synthesized feature statistics, thus introducing an element of uncertainty into the training process.
Furthermore, variations between features hold an inherent semantic value, as indicated by certain generative research studies such as \cite{DBLP:journals/corr/abs-2108-13624} and \cite{DBLP:conf/nips/WangPSZHW19}. By analyzing the variances and directions of these features, we can gauge the likelihood of significant semantic alterations. Motivated by this insight, we present a straightforward yet efficient non-parametric technique for uncertainty estimation. We utilize the variance of the feature statistics to provide guidelines for understanding the magnitude of uncertainty.
\begin{equation}
\Sigma_{\boldsymbol{\mu}_{i}}^2=\frac{1}{B} \sum_{b=1}^B\left(\boldsymbol{\mu}_{b}-\mathbb{E}_B[\boldsymbol{\mu}_{b}]\right)^2,
\end{equation}
\begin{equation}
\Sigma_{\boldsymbol{\sigma}_{i}}^2=\frac{1}{B} \sum_{b=1}^B\left(\boldsymbol{\sigma}_{b}-\mathbb{E}_B[\boldsymbol{\sigma}_{b}]\right)^2,
\end{equation}
where $\Sigma_{\boldsymbol{\mu}_i}$ and $\Sigma_{\boldsymbol{\sigma}_i}$ are represented by the uncertainty estimation of the feature mean $\boldsymbol{\mu}_i$ and the feature standard deviation $\boldsymbol{\sigma}_i$, respectively. $B$ is the mini-batch size. 
The significance of uncertainty estimation magnitudes lies in their ability to indicate the likelihood of future changes in associated characteristics. By assessing and quantifying uncertainty, we can gain valuable insights into the potential variations that may occur. The approach of estimating uncertainty at the mini-batch level is particularly useful as it allows us to capture a suitable and relevant range of variations for each feature. One advantage of using mini-batch uncertainty estimates is that they have minimal impact on the training process of the relation extraction model. This means that the model can continue to learn and improve without being heavily influenced by the uncertainty estimates. However, despite not directly affecting the model's training, these estimates play a crucial role in mimicking a diverse set of probable changes. By encompassing a variety of potential variations, the mini-batch uncertainty estimates provide a comprehensive understanding of the characteristic at hand. They enable us to anticipate and prepare for future changes, helping us make informed decisions and adjustments as needed.
Furthermore, these uncertainty estimates can also assist in robust decision-making. By understanding the possible range of variations, we can assess the potential risks and uncertainties associated with the characteristic. This allows us to devise contingency plans, optimize strategies, and mitigate potential negative impacts.

In conclusion, the consideration of uncertainty estimation magnitudes, particularly at the mini-batch level, offers valuable insights into the likelihood of future changes in characteristics. It provides a means to capture a relevant range of variations without hindering the model's training process. This approach allows us to simulate and anticipate diverse changes, facilitating informed decision-making and robust planning.

\subsection{Uncertainty Relational GNN}

We further design an uncertainty relational graph neural network (UR-GNN) to derive representations from the learned Gaussian distribution, enabling better integration of the uncertainty neighbors and relations.

For the knowledge graphs, the relational graph neural network (RGNN) \cite{DBLP:conf/esws/SchlichtkrullKB18} contributes highly multi-relational data characteristics through relational transformations, relying on the types and direction of relations. It can be expected to reside within the graph encoded through the neighborhood structure for many missing pieces of information and is proved that this method can capture effective relation and neighbor features on the knowledge graph, thus it has been widely used in various tasks of the knowledge graph. The update of RGNN is defined as
\begin{equation}
\boldsymbol{h}_i^{(l+1)}=\operatorname{ReLU} \left(\sum_{r \in \mathcal{R}} \sum_{j \in \mathcal{N}_i^r} \frac{1}{\tilde{d}_{i,r}}\mathbf{W}_r^{(l)} \boldsymbol{h}_j^{(l)} +\mathbf{W}_0^{(l)} \boldsymbol{h}_i^{(l)}\right),
\end{equation}
where $\tilde{d}_{i,r}=\sum_{j \in \mathcal{N}_i^r} \mathbf{A}_{ij}$ is a normalization constant. $\mathbf{A}$ denotes the adjacency matrix. 
The collection of neighbors of entity $i$ under relation $r \in \mathcal{R}$ is denoted by the term $\mathcal{N}_i^r$.

However, after the uncertainty representation, we aggregate the uncertain entity and the relation based on the knowledge graph. 
Current graph convolutions are no longer useful since the feature is parameterized by a probability distribution.
We officially employ an uncertain relational graph neural layer to do convolution operations between Gaussian distributions in the spirit of \cite{DBLP:conf/kdd/ZhuZ0019}. 
Denote $\boldsymbol{h}_i^{(l)}  \sim N\left(\boldsymbol{\mu}_i^{(l)}, \boldsymbol{\sigma}_i^{(l)}\right)$ as the representation of entity $i$ in the $l$-th layer.
Given that all hidden representations of entities must be independent and that the Gaussian distribution \cite{lecam1965distribution} is additive, we aggregate the neighborhoods as follows:
\begin{equation}\label{eq:agg}
\begin{aligned}
\boldsymbol{h}_{\mathcal{N}_i}^{(l)}=&\sum_{r \in \mathcal{R}}\sum_{j \in \mathcal{N}_i^r} \frac{1}{\tilde{d}_{i,r}} \boldsymbol{h}_j^{(l)} \sim \\
& N\left(\sum_{r \in \mathcal{R}}\sum_{j \in \mathcal{N}_i^r} \frac{1}{\tilde{d}_{i,r}} \boldsymbol{\mu}_j^{(l)} , \sum_{r \in \mathcal{R}}\sum_{j \in \mathcal{N}_i^r} \frac{1}{\tilde{d}_{i,r}^2} \boldsymbol{\sigma}_j^{(l)} \right).
\end{aligned}
\end{equation}
We propose a variance-based attention technique that assigns the neighbors different weights due to the varying relevance of the local neighborhood information.
It makes intuitive sense that an entity with a smaller variance is more significant. To manage the impact of weights, we employ a smooth exponential function:
\begin{equation}\label{eq:att}
\alpha_{i,r}^{(l)}=\frac{\exp \left( \operatorname{ReLU} \left(\mathbf{W}_a^{(l)} \boldsymbol{\sigma}_i^{(l)}\right)\right)}{\sum_{j \in \mathcal{N}_i^r}  \exp  \left( \operatorname{ReLU} \left( \mathbf{W}_a^{(l)}  \boldsymbol{\sigma}_j^{(l)} \right) \right)},
\end{equation}
where $\alpha_{i,r}^{(l)}$ stands for attention weights of entity $i$ in the $l$-th layer and $\mathbf{W}_a^{(l)}$ denotes a weight vector followed by a $\operatorname{ReLU}$ activation function. 
This function allows for a gradual transition between different weights and ensures a more balanced representation of the local neighborhood information. By applying the exponential function, we can effectively capture the relative importance of each neighbor entity.
It provides a flexible and adaptive way to assign weights to neighboring entities. By considering the variance of each entity, we can better discriminate between influential and less influential neighbors, thereby incorporating more informative and relevant local neighborhood information.
By incorporating this technique, our model can adaptively focus on significant entities while attenuating the influence of less important ones. This helps to improve the quality and relevance of the learned representations, as it allows for a more fine-grained consideration of the local context.
Eq.(\ref{eq:agg}) can be adjusted as follows when taking the variance-based attention into account:
\begin{equation}
\begin{aligned}
&\boldsymbol{h}_{\mathcal{N}_i}^{(l)}=\sum_{r \in \mathcal{R}}\!\sum_{j \in \mathcal{N}_i^r} \frac{1}{\tilde{d}_{i,r}} \left(\boldsymbol{h}_j^{(l)} \!\cdot\! \alpha_{j,r}^{(l)}\right) \sim \\
&N\!\!\left(\sum_{r \in \mathcal{R}}\!\sum_{j \in \mathcal{N}_i^r} \frac{1}{\tilde{d}_{i,r}}\!\!\left(\!\boldsymbol{\mu}_j^{(l)}\!\cdot\! \alpha_{j,r}^{(l)}\right)\!, \!\sum_{r \in \mathcal{R}}\!\sum_{j \in \mathcal{N}_i^r}\! \frac{1}{\tilde{d}_{i,r}^2} \!\!\left(\!\boldsymbol{\sigma}_j^{(l)} \!\cdot\! \alpha_{j,r}^{(l)} \!\cdot\! \alpha_{j,r}^{(l)}\!\right)\!\!\right)\!,
\end{aligned}
\end{equation}
where the element-wise product operation is indicated by $[\cdot]$. 
To ensure the proper integration of neighbor and relation information, our approach applies attention weights independently for each dimension. This independent weighting enables us to effectively combine the different aspects of information and prioritize them accordingly.
By applying attention weights independently, we can accurately capture the relevance and significance of each dimension in the feature space. This allows us to extract and integrate relevant information from both neighbor entities and relations in a more comprehensive manner.

We use non-linear activation functions and learnable filters to $\boldsymbol{h}_{\mathcal{N}_i^r}^{(l)}$ aiming to acquiring $\boldsymbol{h}_{i}^{(l+1)}$. While $\boldsymbol{h}_{\mathcal{N}_i^r}^{(l)}$ has a Gaussian distribution, computing $\boldsymbol{h}_{i}^{(l+1)}$ is theoretically impossible. In this case, we immediately apply non-linear activation functions to variances and means, respectively, and layer-specific parameters to the means. Consequently, the following definition applies to the unknown relational graph convolution.
\begin{equation}
\begin{aligned}
&\boldsymbol{\mu}_{i}^{(l+1)}=\operatorname{ReLU}\left(\sum_{r \in \mathcal{R}}\sum_{j \in \mathcal{N}_i^r} \frac{1}{\tilde{d}_{i,r}} \mathbf{W}_{\mu,r}^{(l)} \left(\boldsymbol{\mu}_{j}^{(l)} \cdot \alpha_{j,r}^{(l)}\right) \right), \\
&\boldsymbol{\sigma}_{i}^{(l+1)}=\operatorname{ReLU}\left(\sum_{r \in \mathcal{R}}\sum_{j \in \mathcal{N}_i^r} \frac{1}{\tilde{d}_{i,r}^2} \mathbf{W}_{\sigma,r}^{(l)} \left(\boldsymbol{\sigma}_{j}^{(l)} \cdot \alpha_{j,r}^{(l)} \cdot \alpha_{j,r}^{(l)}\right) \right),
\end{aligned}
\end{equation}
where $\mathbf{W}_{\mu,r}^{(l)},\mathbf{W}_{\sigma,r}^{(l)}$ are learnable transformation matrices. The updates of entity representations in the ($l+1$)-th layer are the integration of neighbors' representations in the $l$-th layer, which learns uncertainty entity representation with the help of neighborhood. Different from RGCN, the relation representation relies on multiple entity pairs of the specific relation in the few-shot situation. 
\begin{equation}
\begin{aligned}
&\boldsymbol{\mu}_{r}^{(l+1)}=\operatorname{ReLU}\left(\sum_{i \in \mathcal{S}_r} \frac{1}{|\mathcal{S}_r|} \mathbf{W}_{\mu}^{(l)} \left(\boldsymbol{\mu}_{i}^{(l)} \cdot \alpha_{i,\mathcal{S}_r}^{(l)}\right)\right), \\
&\boldsymbol{\sigma}_{r}^{(l+1)}=\operatorname{ReLU}\left(\sum_{i \in \mathcal{S}_r} \frac{1}{|\mathcal{S}_r|} \mathbf{W}_{\sigma}^{(l)} \left(\boldsymbol{\sigma}_{i}^{(l)} \cdot \alpha_{i,r}^{(l)} \cdot \alpha_{i,\mathcal{S}_r}^{(l)}\right) \right),
\end{aligned}
\end{equation}
where $\mathbf{W}_{r,\mu}^{(l)},\mathbf{W}_{r,\sigma}^{(l)}$ are learnable transformation matrices. $S_r$ is the support set of relation $r$. Simlar with Eq.(\ref{eq:att}), $\alpha_{i,\mathcal{S}_r}^{(l)}$ is the variance weight in $\mathcal{S}_r$ using softmax normalization. The updates of relations representations in the ($l+1$)-th layer are the integration of entity representations in the $l$-th layer, which are connected to the relation.

\subsection{Uncertainty Optimization}
In order to make predictions with only a few samples, we design the uncertain completion prediction utilizing multiple sampling from the learned distributions to measure the completion score. Additionally, to  leverage the uncertainty in the probabilistic representations and to be robust to the noises in few-shot scenarios, we propose the uncertainty joint loss.

\subsubsection{Uncertain Completion Prediction}
It is possible to establish the Gaussian distribution for the probabilistic feature statistics after the uncertainty representation module of each feature channel has been determined. To further leverage the uncertainty in the probabilistic representations, we employ the random sampling approach to describe the uncertainty.
Here, we use the re-parameterization method \cite{DBLP:journals/corr/KingmaW13} to make the sampling operation differentiable. 
The corresponding distributions can be used to generate the final entity embedding at random:
\begin{equation}
\begin{aligned}
\boldsymbol{z}_i=\boldsymbol{\beta_{i}}+\epsilon_z \boldsymbol{\gamma_{i}}=(\boldsymbol{\mu_{i}}+\epsilon_\mu \Sigma_{\boldsymbol{\mu_{i}}}) + \epsilon_z (\boldsymbol{\sigma_{i}}+\epsilon_\sigma \Sigma_{\boldsymbol{\sigma_{i}}}),
\end{aligned}
\label{zi}
\end{equation}
where the typical Gaussian distribution is followed by the variables $\epsilon_\mu$, $\epsilon_\sigma$ and $\epsilon_z$. 
The utilization of random sampling within our framework plays a pivotal role in enriching the pool of feature statistics information. This enrichment is achieved by harnessing the power of the Gaussian distribution, which inherently provides a spectrum of diverse combinations of orientations and intensities. The benefits of this approach extend to the training of our model, especially in scenarios where feature statistics exhibit ambiguity.
This resilience to potential statistics changes and the resulting improved generalization ability are invaluable assets, making our model more adaptable and reliable in real-world scenarios, where data can be dynamic and ever-changing.

To calculate the semantic similarity for tail entity prediction, we build a metric function $P{\left(\boldsymbol{z}_h, \boldsymbol{z}_t\right)}=||\boldsymbol{z}_{h}\boldsymbol{z}_{r}-\boldsymbol{z}_t||$, where $\boldsymbol{z}_h$, $\boldsymbol{z}_r$ and $\boldsymbol{z}_t$ is the representation of the head entity, relation and tail entity calculated by Equation~\ref{zi}. It becomes clear that when there is uncertainty around a critical context in KG, the gap can be significant. 
We random sample $\boldsymbol{z}_i$ embedding $m$ times for each query as augmented instances. The metric function of the $k$-th augmentation $P{\left(\boldsymbol{z}_h, \boldsymbol{z}_t, k\right)}=||\boldsymbol{z}_{h}^k\boldsymbol{z}_{r}^k-\boldsymbol{z}_t^k||$, where $\boldsymbol{z}_{h}^k$, $\boldsymbol{z}_{r}^k$ and $\boldsymbol{z}_t^k$ mean the k-th sample of $\boldsymbol{z}_{h}$, $\boldsymbol{z}_{r}$ and $\boldsymbol{z}_t$. The average scores $\bar{P}{\left(\boldsymbol{z}_h, \boldsymbol{z}_t\right)}$ is calculated as:
\begin{equation}
\begin{aligned}
\bar{P}{\left(\boldsymbol{z}_h, \boldsymbol{z}_t\right)}=\frac{\sum_{k=1}^m P{\left(\boldsymbol{z}_h, \boldsymbol{z}_t, k\right)}}{m},
\end{aligned}
\end{equation}
where $m$ is the number of random samples. The average score $\bar{P}{\left(\boldsymbol{z}_h, \boldsymbol{z}_t\right)}$ calculates the average score to learn a more robust representation suit for the uncertainty model training.

\subsubsection{Uncertainty Joint Loss}
We design an uncertainty joint loss function, which utilizes the entity completion loss, the uncertain mutual information losses of entity classification scores, and the KL-divergence loss to evaluate the model performance for the multi-perspective assessment.

\textbf{Completion Loss.} The completion loss is the sum of average scores of positive and negative entity pairs in the query set:
\begin{equation}
\mathcal{L}_{com}=\sum_{\left(\boldsymbol{z}_{h}, \boldsymbol{z}_t\right) \in \mathcal{Q}_r}\left[\delta+\bar{P}{\left(\boldsymbol{z}_h, \boldsymbol{z}_t\right)}-\bar{P}{\left(\boldsymbol{z}_h, \boldsymbol{z}_t^{\prime}\right)}\right]_{+},
\end{equation}
where $[x]_{+}$ denotes the positive part of $x$, $z_{t}$ is the possible entities of $T$, $A_{k}$ is the augmentations of the $k$-th representation, and $\delta$ represents the margin.
$\bar{P}{\left(\boldsymbol{z}_h, \boldsymbol{z}_t\right)}$ is the score for the negative entity pair $(\boldsymbol{z}_h, \boldsymbol{z}_t^{\prime})$ corresponding to the current positive entity pair $(\boldsymbol{z}_h, \boldsymbol{z}_t)\in \mathcal{Q}_r$, where $(\boldsymbol{z}_h, r, \boldsymbol{z}_t^{\prime})\notin \mathcal{Q}_r$.

\textbf{Uncertainty Mutual Information (UMI) Loss.} To measure the uncertainty of the prediction and calculate the average prediction scores for each query instance, we propose mutual uncertainty information loss. 
We use the tractable method suggested by \cite{DBLP:journals/corr/abs-1112-5745}, which roughly approximates the inferred mutual information by varying the model's parameters to produce different predictions, computing mutual information. In this research, we suggest using a variety of data-augmentation techniques to calculate the uncertainty of model prediction outcomes. In particular, the Shannon entropy $H(\cdot)$ is used to calculate the mutual uncertainty information loss $\mathcal{L}_{\text{UMI}}$ based on the classification scores of the enhanced cases $m$:
\begin{equation}
\mathcal{L}_{\text{UMI}}=H\left(\bar{P}{\left(\boldsymbol{z}_h, \boldsymbol{z}_t\right)}\right)-\frac{1}{m}\sum_{k=1}^m H\left({P}{\left(\boldsymbol{z}_h, \boldsymbol{z}_t,k\right)}\right),
\end{equation}
where $m$ is the number of random samples. It refines representations based on uncertainty information provided by query instances. 
By incorporating mutual uncertainty information loss into our analysis, we can not only measure the uncertainty of individual predictions but also obtain an average prediction score for each query instance. 

\textbf{KL-divergence (KL) Loss.} The UR-GNN is calculated by using the calculation rule of Gaussian distribution. To ensure that the representations are Gaussian distributions, the entity and relation distributions should be characterized similarly to $N(\mathbf{0}, \boldsymbol{I})$. Thus, we also provide a regulation loss to enforce entity representations as Gaussian distributions:
\begin{equation}
\mathcal{L}_\text{reg}=\operatorname{KL}\left(N(\boldsymbol{\mu}_i^{(l)}, \boldsymbol{\sigma}_i^{(l)}) \| N(\mathbf{0}, \boldsymbol{I})\right),
\end{equation}
where $N(\boldsymbol{\mu}_i^{(l)}, \boldsymbol{\sigma}_i^{(l)})$ is the Gaussian distributions for all entities in the $l$-th layer, and $\operatorname{KL}$ is the KL-divergence \cite{kullback1951information}.

By leveraging uncertainty mutual information loss $\mathcal{L}_\text{UMI}$ of the reference set aggregator and the KL-divergence loss $\mathcal{L}_\text{reg}$, the final objective function is defined as follows.
\begin{equation}
\mathcal{L}_{\text{joint}}=\mathcal{L}_{com}+\lambda_{1} \mathcal{L}_{\text{UMI}}+\lambda_{2} \mathcal{L}_\text{reg},
\end{equation}
where $\lambda_{1}$ and $\lambda_{2}$ are the trade-off factors between prediction loss $\mathcal{L}_{com}$, UMI loss $\mathcal{L}_{\text{UMI}}$ and KL-divergence loss $\mathcal{L}_\text{reg}$. We optimize all training documents in the mini-batch strategy.

\section{Experiment}
To assess the effectiveness of our model, we run link prediction experiments and provide discussions on the model variants.

\subsection{Datasets and Evaluation Metrics}
\textbf{Datasets.} Two public benchmark datasets, NELL \cite{mitchell2018never} and Wiki \cite{vrandevcic2014wikidata}, are used in this paper. 
The NELL is a comprehensive KG dataset learning continuously over time, including wide-ranging types of knowledge, while Wiki is a dataset based on Wikipedia. 
Following GMatching\cite{xiong-etal-2018-one}, relations in both datasets with fewer than $500$ but more than $50$ triples are chosen to create few-shot tasks. There are $67$ and $183$ few-shot relations in the NELL and Wiki datasets, with $51/5/11$ and $133/16/34$ used as the partition of training, validating, and testing, respectively. Additionally, both datasets include candidate entities built using the constraint of entity types \cite{xiong-etal-2018-one} for each connection.

\textbf{Evaluation Metrics.}
The effectiveness of per test triple's ranking among all potential tail entity replacements is assessed using the following metric. On both datasets, we present two common evaluation metrics: Hits@N and MRR. Hits@N is the percentage of accurate entities rated in the top $N$, with $N = 1, 5, 10$, and the MRR metric is the mean reciprocal rank.

\subsection{Comparision Methods}
Our approach is compared with the following two sets of baselines to assess its effectiveness:

\textbf{KG Embedding Methods.} 
These models propose to map relations and entities to a vector space considering specific constraints, i.e., how to model relational structures in KGs. The entity/relation embeddings carry their inherent structural features and can achieve satisfactory results with sufficient training samples. We employ four popular KG embedding techniques as our baselines: 

(1) TransE~\cite{bordes2013translating} interprets a relation as a translation process based on additive operation.

(2) DistMult \cite{DBLP:journals/corr/YangYHGD14a} adopts simple bilinear transformation between entity vectors based on multiplicative operation. 

(3) ComplEx \cite{DBLP:conf/icml/TrouillonWRGB16} follows the formulation in DistMult except for mapping entities and relations to the complex vector space, modeling structural information as multiplications between different parts of complex vectors.

(4)~RotatE \cite{DBLP:conf/iclr/SunDNT19} defines relations as rotation operations between head and tail entities in the complex vector space.
Every KG embedding approach needs enough training triples for every relation and teaches static knowledge graph representations.

\begin{table*}[t]
\caption{Main experiments of 5-shot entity completion (\%). \textbf{Bold}: best results, Underlined: second-best results, "$\Delta$ Avg": average decrease value of four metrics compared to our model. "$\downarrow$": average decrease compare to UFKGC. "$\uparrow$": improvement compared to the second-best result. "–": results are not available. 
}
\centering
\renewcommand\arraystretch{1.3}
\resizebox{\linewidth}{!}{
\begin{tabular}{l|ccccc|ccccc}
\toprule
  \multirow{2}*{\textbf{Models}} &  & & \textbf{NELL dataset}  &    &   &  & & \textbf{Wiki dataset}  &    &   \\\cline{2-6}\cline{7-11}
     & MRR  &  Hits@10    &   Hits@5   &  Hits@1  &  $\Delta$Avg  &   MRR  &  Hits@10    &   Hits@5   &  Hits@1 & $\Delta$Avg  \\
\midrule
 \textbf{ TransE \cite{bordes2013translating} }   & 17.4  & 31.3  & 23.1  & 10.1 & $\downarrow$25.5  & 13.3  & 18.7  & 15.7  & 10.0 &  $\downarrow$31.5  \\  
  \textbf{ DistMult \cite{DBLP:journals/corr/YangYHGD14a} }   & 20.0  & 31.1  & 25.1  & 13.7 & $\downarrow$23.5  & 7.1  & 15.1  & 9.9  & 2.4 &  $\downarrow$37.3 \\
 \textbf{ ComplEx \cite{DBLP:conf/icml/TrouillonWRGB16}}   & 18.4  & 29.7  & 22.9  & 11.8  &  $\downarrow$25.3   &  8.0  & 18.1  & 12.2  & 3.2  & $\downarrow$35.6   \\ 
 \textbf{ RotatE \cite{DBLP:conf/iclr/SunDNT19} }   &  17.6 & 32.9 & 24.7 & 10.1 &  $\downarrow$24.7  &  4.9 & 9.0 & 6.4 & 2.6 & $\downarrow$40.2   \\  \midrule
  \textbf{ GMatching \cite{xiong-etal-2018-one} }    &  17.6  & 29.4 & 23.3 & 11.0 &  $\downarrow$25.7  & 26.3  & 38.7 & 33.7 & 19.7 & $\downarrow$16.3  \\ 
  \textbf{  MetaR \cite{chen-etal-2019-meta}}    &  20.9  & 35.5 & 28.0 & 14.1  & $\downarrow$21.4   &  32.3  & 41.8 & 38.5 & 27.0 &   $\downarrow$11.0 \\
 \textbf{  FSRL \cite{DBLP:conf/aaai/ZhangYHJLC20} }    &  15.3  & 31.9 & 21.2 & 7.3 & $\downarrow$27.1  &  15.8  & 28.7 & 20.6 & 9.7 & $\downarrow$27.2   \\
   \textbf{ FAAN \cite{DBLP:conf/emnlp/ShengGCYWLX20} }    &  27.9  & 42.8 & 36.4 & 20.0 & $\downarrow$14.2   &  34.1  & 46.3 & 39.5 & 28.1 &  $\downarrow$8.9  \\
    \textbf{  GANA \cite{DBLP:conf/sigir/NiuLTGDLWSHS21} }    &  34.4  & \underline{51.7} & 43.7 & 24.6 & $\downarrow$7.4   &  35.1  & 44.6 & 40.7 & 29.9 & $\downarrow$8.4  \\ 
    \textbf{ P-INT \cite{DBLP:conf/emnlp/XuZKDCLL21} }    & \underline{40.5}   & 50.6 & \underline{50.3} & \underline{31.7} & $\downarrow$2.7    &  - & - & - & - & -  \\
    \textbf{ YANA \cite{DBLP:conf/ksem/LiangZCYY22} }    & 29.4   & 42.1 & 36.4 & 23.0 & $\downarrow$13.3  &  \underline{38.0} & \underline{52.3} & \underline{44.2} & \underline{32.7} & $\downarrow$4.1  \\
    \midrule
    \textbf{$\text{ UFKGC (Ours) }$}    &  \textbf{41.7} ($\uparrow$1.2)  & \textbf{58.8} ($\uparrow$7.1)  & \textbf{51.1} ($\uparrow$0.8) & \textbf{32.4} ($\uparrow$0.7) & -  &  \textbf{43.1} ($\uparrow$5.1)  & \textbf{54.0} ($\uparrow$1.7) & \textbf{49.1} ($\uparrow$4.9) & \textbf{37.5} ($\uparrow$4.8) & -  \\ 
\bottomrule
\end{tabular}
}
\label{main}
\end{table*}

\begin{table*}[t]
\caption{Variant experiments (\%). "w/o": removing the corresponding module from the complete model.
"$\Delta$ Avg": average decrease value of four metrics compared to our model. "$\downarrow$": average decrease compare to UFKGC. UR: Uncertainty Representation.}
\centering
\renewcommand\arraystretch{1.3}
\resizebox{\linewidth}{!}{
\begin{tabular}{l|ccccc|ccccc}
\toprule
  \multirow{2}*{\textbf{Variants}}&  & & \textbf{NELL dataset}  &    &   & & &  \textbf{Wiki dataset}  &    &   \\\cline{2-6}\cline{7-11}
    &  MRR &  Hits@10   &   Hits@5    &  Hits@1  &   $\Delta$Avg   &    MRR  &  Hits@10    &   Hits@5   &  Hits@1  &  $\Delta$Avg  \\
\midrule
 \textbf{$\text{UFKGC (Ours)}$} &  \textbf{41.7}  & \textbf{58.8}  & \textbf{51.1} & \textbf{32.4} & - &  \textbf{43.1}  & \textbf{54.0} & \textbf{49.1} & \textbf{37.5} &  - \\ \midrule
 {  }\textbf{ w/o UR}   &  35.6  & 53.1 & 46.5 & 27.4 &  $\downarrow$5.4  &  38.3  & 47.8 & 45.1 & 32.7  &   $\downarrow$5.0  \\  
  {  }\textbf{ w/o Uncertainty Estimation}   &  37.9  & 55.6 & 47.8 & 30.3 &   $\downarrow$3.1 &  41.2  & 49.2 & 45.9 & 35.0 &  $\downarrow$3.1  \\  
{  }\textbf{ w/o Uncertainty Attention}   &  37.3  & 57.1 & 46.2 & 30.3 & $\downarrow$3.3 &  41.5  & 49.1 & 47.7 & 37.0 & $\downarrow$2.1 \\
 {  }\textbf{ w/o UMI loss}  &  39.6  & 56.4 & 48.2 & 30.5 & $\downarrow$2.3 &  41.9  & 49.9 & 47.3 & 36.8 &  $\downarrow$2.0 \\ 
  {  }\textbf{ w/o KL loss}  &  39.4  & 56.2 & 48.4 & 30.1 &  $\downarrow$2.5  &  41.6  & 49.7 & 46.9 & 36.2 & $\downarrow$2.3  \\
    {  }\textbf{ w/o UMI and KL losses}  &  33.2  & 52.9 & 44.7 & 29.5 &  $\downarrow$5.9  &  39.5  & 47.2 & 45.1 & 33.6 & $\downarrow$4.6  \\     
 
\bottomrule
\end{tabular}
}
\label{Ablation}
\end{table*}

\textbf{Few-shot KG Completion Methods.} 
These models focus on the few-shot scenario in KG completion by integrating metric-based or optimization-based meta-learning approaches with pre-trained embeddings, generally taking advantage of local structures and relational semantic information to generate robust embeddings for relations or entity pairs for further usage, and are able to deliver state-of-the-art performances on the NELL and Wiki datasets:

(1) In GMatching (MaxP) \cite{xiong-etal-2018-one}, a matching network and a neighbor encoder are used, however, it is assumed that each neighborhood makes a similar contribution. 

(2) MetaR \cite{chen-etal-2019-meta} creates prediction by transferring the shared data from the reference to the queries depending on a special optimization method. 

(3) FSRL \cite{DBLP:conf/aaai/ZhangYHJLC20} encodes neighbors using a fixed attention approach and uses a recurrent autoencoder for aggregating references.

(4) FAAN~\cite{DBLP:conf/emnlp/ShengGCYWLX20} proposes a dynamic attention mechanism in entity completion of a small sample to learn the multi-semantic task relationship.

(5) GANA~\cite{DBLP:conf/sigir/NiuLTGDLWSHS21} proposes a gating mechanism to obtain contextual semantic information on the neighborhood task relationship and neighborhood information.

(6) YANA \cite{DBLP:conf/ksem/LiangZCYY22} constructs a local pattern graph with support and query set entities and their relationships to learn more robust representations for solitary entities.

(7) P-INT \cite{DBLP:conf/emnlp/XuZKDCLL21} leverages the directed subgraph structure between head and tail entities to represent an entity pair and computes the interaction between paths from support and query set for similarity scores.

Each of the aforementioned techniques only teaches static representations of entity or reference, omitting dynamic features. In contrast to these methods, our method transfers representation into Gaussian distribution to better model the uncertainty of features and learns more robust entity representations via the UR-GNN with an uncertainty joint loss.

\subsection{Implementation Details}
For all baselines, we adopt the best hyper-parameters and copy existing results reported in the literature~\cite{xiong-etal-2018-one, chen-etal-2019-meta, DBLP:conf/aaai/ZhangYHJLC20, DBLP:conf/emnlp/ShengGCYWLX20, DBLP:conf/sigir/NiuLTGDLWSHS21, DBLP:conf/emnlp/XuZKDCLL21, DBLP:conf/ksem/LiangZCYY22}. 

\textbf{Experimental Settings.}
Our implementation is based on PyTorch\footnote{https://pytorch.org/} to develop the few-shot KG completion. All experiments were conducted on a server with one GPU (Tesla V100). 
Note that all the hyper-parameter settings are tuned on the validation set by the grid search with $5$ trials to their optimum values. The early stop strategy is selected to prevent the model from overfitting. We put each approach through a 5-shot knowledge graph completion assignment. 
In order to train models, we utilize all triples on the background knowledge graph and training sets along with a limited number of reference triples from the validation and testing sets. We initialize entity embeddings via TransE.

\textbf{Hyperparameters of UFKGC.}
Prior to model training, the entity neighbors are fixed at a maximum of $50$ on both datasets and are randomly picked and fixed. 
The learning rate by the Adam optimizer is initialized at $4e-5$ and $5e-5$ on the NELL and Wiki datasets. 
The batch size is $512$, and the dropout rate is $0.1$ and $0.3$ for the two datasets in order to prevent over-fitting. We use AdamW~\cite{DBLP:conf/iclr/LoshchilovH19} to optimize the parameters. 
For hyper-parameters, the best coefficients $ \lambda_{1}, \lambda_{2}$ are $0.5$ and $0.3$. For the learning rate, we adopt the method of grid search with a step size of $0.0001$. 
To ensure fairness, all baselines use the same dataset partitioning and entity representation dimension, which is set to $128$ dimensions.

\subsection{Main Results}


To verify the effectiveness of our model, we report overall average results in
Table~\ref{main}. It shows performance comparisons on the NELL and Wiki datasets of the 5-shot link prediction task. All the results of these models are obtained from the original papers. 
Note that the results on the Wiki dataset for the few-shot knowledge graph completion are unprovided for P-INT because P-INT is not suitable for sparse dataset scenarios.

\begin{figure*}[!htp]
    \centering
 \subfigure[Impact on the NELL dataset.]{
 \includegraphics[width=\linewidth]{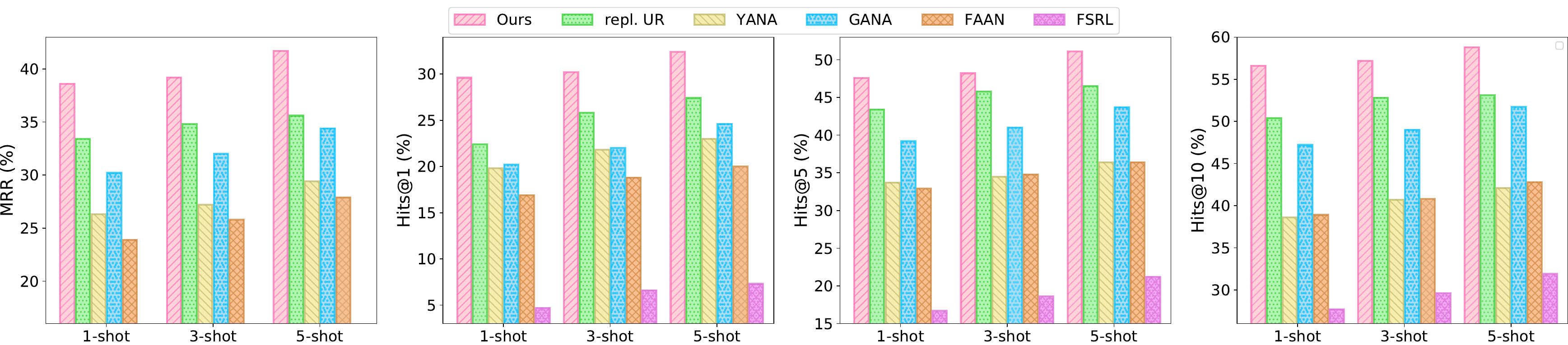}}
 \centering
    \subfigure[Impact on the Wiki dataset.]{\includegraphics[width=\linewidth]{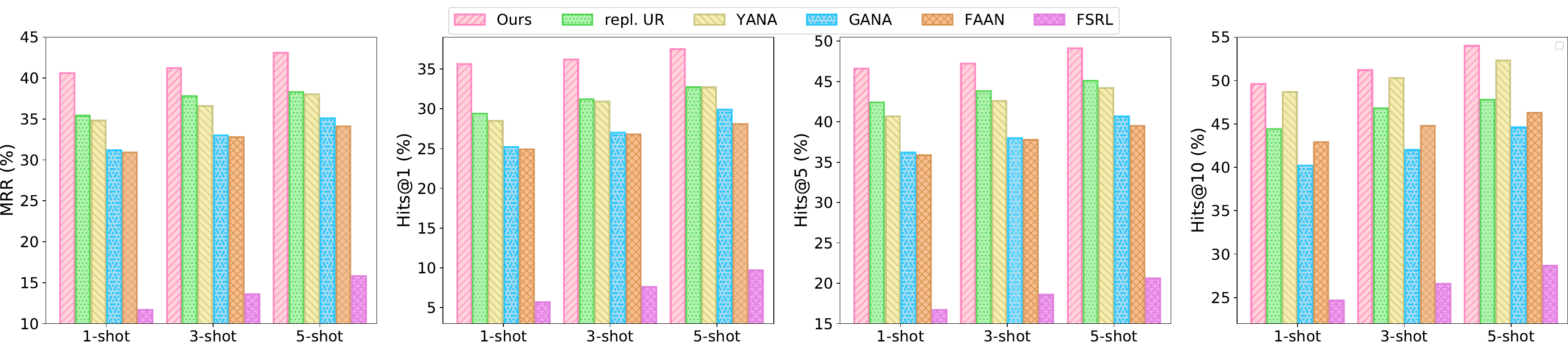}}
    \caption{Discussions for different few-shot situations, compared to the partial outstanding few-shot completion baselines.}
    \label{KGC_Fshot_nell}
\end{figure*}



From the table, we can observe that: 1) Our model outperforms all the baselines of the few-shot knowledge graph completion, in terms of four metrics. It achieves at least 2.7\% and 4.1\% improvements on average on the NELL and Wiki datasets.
Especially, our model improves by at least 5.1\% in the MRR metric on the Wiki dataset and 7.1\% in the Hits@10 on the NELL dataset, respectively. 
It demonstrates that our model learns a more robust representation of entities and relations through integrating the uncertainty information, achieving reliable performance on the few-shot knowledge graph completion task. 
2) Our model outperforms the knowledge graph embedding baselines by an average of $23.5\%$ and $31.5\%$ on the NELL and Wiki datasets in the context of few-shot knowledge graph completion. This signifies the efficacy of incorporating the uncertainty of entities and relations, which proves to be more appropriate for this specific task.
In comparison to existing baselines, our model shows substantial improvement in its ability to complete knowledge graphs with limited data. By considering the uncertainty associated with both entities and relations, our model captures a more comprehensive understanding of the data, leading to more accurate and robust results.
3) Our model outperforms the knowledge graph embedding baselines by an average of $23.5\%$ and $31.5\%$ on the NELL and Wiki datasets in the context of few-shot knowledge graph completion. This signifies the efficacy of incorporating the uncertainty of entities and relations, which proves to be more appropriate for this specific task.
In comparison to existing baselines, our model shows substantial improvement in its ability to complete knowledge graphs with limited data. By considering the uncertainty associated with both entities and relations, our model captures a more comprehensive understanding of the data, leading to more accurate and robust results.
4) Compared to the best performance baseline GANA on the NELL dataset for the few-shot knowledge graph completion, our model improves by an average of 8.4\%. Compared to the best performance baseline YANA on the NELL and Wiki datasets, our model improves by an average of 13.3\% and 4.1\%. The main reason is that our method learns the dynamic features and evaluates the uncertainty for promoting the exploitation of limited samples for entity completion in few-shot optimization.
All observations demonstrate the effectiveness of our model for the few-shot knowledge graph completion task, which models the uncertainty of entities and triples to capture the similarity and differences.

\subsection{Discussion for Model Variants}

To investigate the effectiveness of each module in our uncertainty-aware few-shot knowledge graph completion model, we conduct variant experiments in Table~\ref{Ablation}. It shows comparisons of our whole few-shot knowledge graph completion model with removing some core modules.

From the table, we can observe that: 
1) The impact of the uncertainty representation tends to be more significant. We believe that the reason is that our model reflects the uncertainty of entities and relations, promoting the exploitation of limited samples in a few-shot knowledge graph completion. 2) When the uncertainty estimation module is removed, the performance decreases, suggesting that measuring uncertainty contributes to understanding the implied dynamic features. 
3) The "Uncertainty Attention" component achieves a positive impact on the results. "w/o Uncertainty Attention" removes the variance-based uncertainty attention weights. This component helps to discern valuable entity features with higher certainty, demonstrating that the uncertainty attention mechanism helps to learn a better representation for the few-shot knowledge graph completion.
4) By removing the Uncertain Mutual Information (UMI) loss and the KL-divergence (KL) loss jointly, the model performance decreases dramatically. It illustrates the effectiveness of the design of join loss, which enhances the robustness of entity representation and minimizes the worst-case gap for different relations of the entity completion task. 5) By removing the UMI loss, the performance basically decreased on all metrics. It demonstrates that the uncertain mutual information loss module effectively estimates the uncertainty of the few-shot knowledge graph completion scores for each query instance.  
All the observations demonstrate the effectiveness of each component in our model.

\begin{table*}[t]
\caption{UR-GNN replacement experiments (\%). "repl.": replacing the corresponding module with the other module.}
\centering
\renewcommand\arraystretch{1.3}
\resizebox{\linewidth}{!}{
\begin{tabular}{l|ccccc|ccccc}
\toprule
  \multirow{2}*{\textbf{Variants}}&  & & \textbf{NELL dataset}  &    &   & & &  \textbf{Wiki dataset}  &    &   \\\cline{2-6}\cline{7-11}
    &  MRR  &  Hits@10    &   Hits@5    &  Hits@1  &   $\Delta$ Avg  &    MRR &  Hits@10    &   Hits@5    &  Hits@1   &  $\Delta$ Avg    \\
\midrule
 \textbf{$\text{UFKGC (Ours)}$} &  \textbf{41.7}  & \textbf{58.8}  & \textbf{51.1} & \textbf{32.4} & - &  \textbf{43.1}  & \textbf{54.0} & \textbf{49.1} & \textbf{37.5} &  -  \\ \midrule
     {  }\textbf{ repl. RGCN}  &  38.0  & 57.5 & 46.7 & 31.1 & $\downarrow$2.7  &  42.3  & 49.8 & 47.5 & 37.6 & $\downarrow$1.6  \\ 
   {  }\textbf{ repl. GCN}  &  35.7  & 58.2 & 45.8 & 30.4 & $\downarrow$3.5  &  40.5  & 47.3 & 47.0 & 37.0 &  $\downarrow$3.0  \\ 
   {  }\textbf{ repl. TransE}  &  21.4  & 36.8 & 27.9 & 27.4 & $\downarrow$17.6  &  20.1  & 26.3 & 22.5 & 18.7 &  $\downarrow$24.0  \\ 
   {  }\textbf{ repl. DistMult}  &  28.5  & 39.2 & 30.0 & 24.7 & $\downarrow$15.4  &  24.8  & 27.4 & 28.3 & 16.2 &  $\downarrow$21.8  \\ 
       {  }\textbf{ repl. ComplEx}  &  27.3  & 37.9 & 29.1 & 23.5 & $\downarrow$16.6  &  23.0  & 29.5 & 29.4 & 14.3 &  $\downarrow$21.9  \\ 
    {  }\textbf{ repl. RotatE}  &  30.9  & 46.7 & 30.3 & 26.8 & $\downarrow$12.3  &  24.4  & 32.0 & 34.6 & 17.1 &  $\downarrow$18.9 \\ 

\bottomrule
\end{tabular}
}
\label{exp:UR-GNN}
\end{table*}

\begin{figure*}[!htp]
    \centering
 \subfigure[Impact on NELL dataset.]{
     \includegraphics[width=0.44\linewidth]{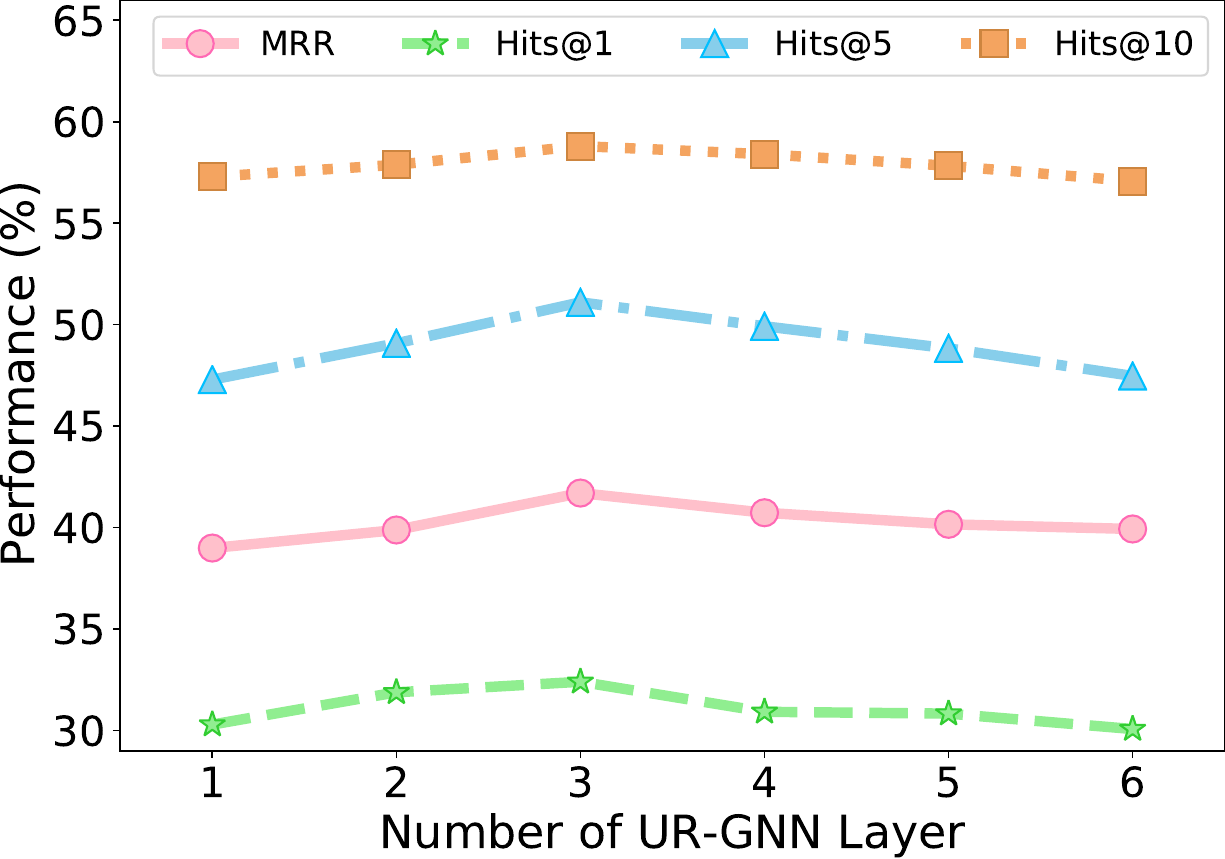}}
     \hspace{10mm} 
    \centering  
 \subfigure[Impact on the Wiki dataset.]{
 \includegraphics[width=0.44\linewidth]{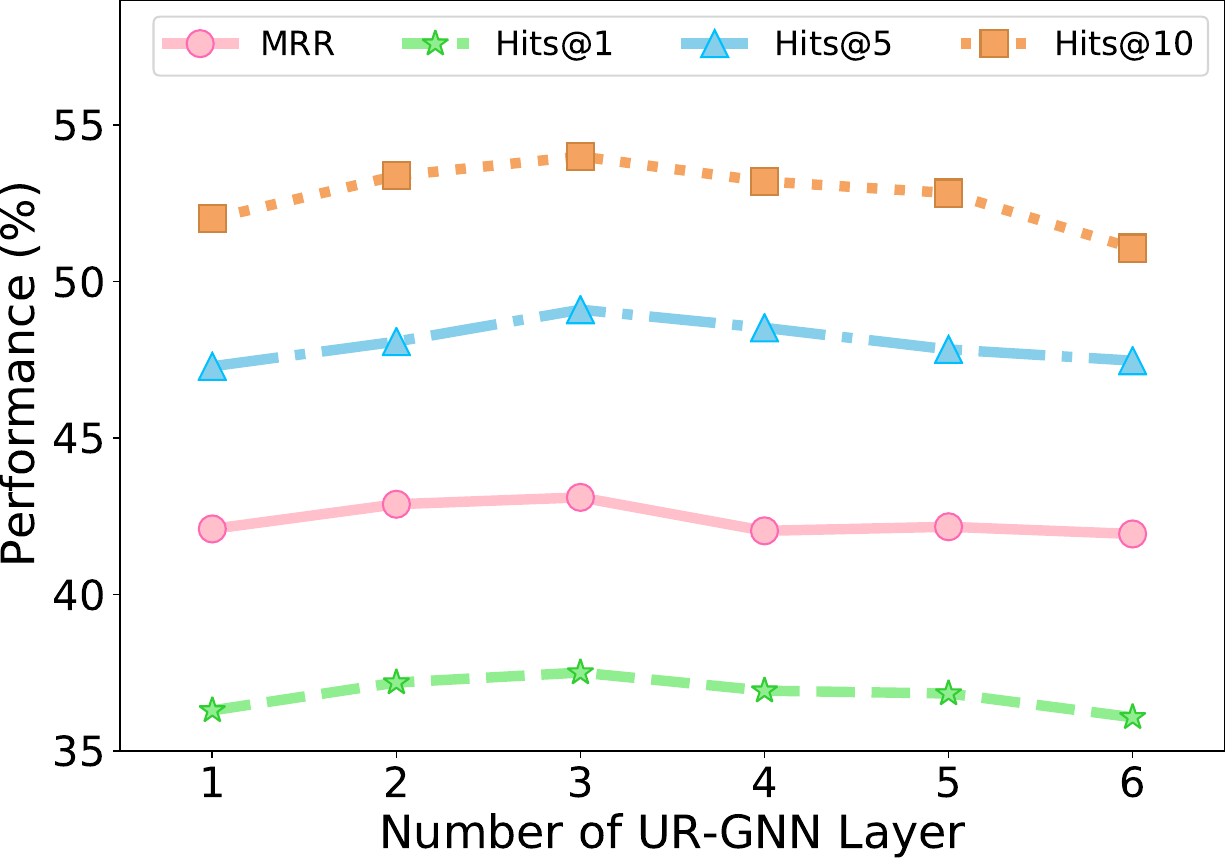}}
    \caption{Impacts of the UR-GNN layer on the NELL and Wiki dataset for the 5-shot situation.}
    \label{KGC_layer}
\end{figure*}



\subsection{Discussions on Uncertainty Representation}

To further investigate the impact of the uncertainty representation module, we report results by the 1-shot, 3-shot, and 5-shot situations for the NELL and Wiki datasets on the MRR, Hits@1, Hits@5, and Hits@10 metrics on the few-shot knowledge graph completion task, as shown in Figure~\ref{KGC_Fshot_nell}. We compare not only with the partially outstanding few-shot knowledge graph completion baseline approaches but also with our approach without the uncertainty representation module.

From the figure, we can observe that: 1) The variants without uncertainty representation "repl. UR" module significantly decline on all evaluation metrics and situations, which demonstrates that our model improves more dramatically when we model the uncertainty of entities and relations. 2) Our model is more affected by the 1-shot knowledge graph completion task compared to other baselines. The reason we think this is that our model exploits the uncertainty to promote the exploitation of limited samples in few-shot optimization and learn a robust representation. 3) Our model performs best in different few-shot situations, demonstrating that our method is appropriate for few-shot completion tasks by considering the uncertainty of features and estimating the uncertainty. 4) Our model renders similar trends on the two benchmarks, reflecting that our model is applicable in different completion circumstances. It demonstrates that our model makes full use of semantic information and uncertainty modeling promoting the exploitation of limited samples in the few-shot optimization.
All the observations demonstrate that our model makes more effective use of uncertainty representation as well as considering modeling the uncertainty of features for the few-shot knowledge graph completion task.

\subsection{Discussion on UR-GNN }
\label{sec:Layer}
To further investigate the impact of each component in the UR-GNN, we conducted variants on the proposed UR-GNN. The results are as shown in Table~\ref{exp:UR-GNN}.


This table provides an overview of the performance of different variants of the UR-GNN, allowing us to analyze the contribution of each component towards the overall performance. Upon analyzing the results presented in the table, the following observations can be made:
1) It is evident that the proposed UR-GNN model outperforms previous typical models in the few-shot knowledge graph completion task. This finding highlights the benefits of incorporating an uncertain manner into the model, as it enables the UR-GNN to capture valuable information for completion. The introduction of uncertainty in modeling entities and relations allows for a more comprehensive and robust representation, leading to improved performance.
2) The performance of the UR-GNN model remains superior when compared to alternative GNN-based models. Replacing the UR-GNN module with these models results in a decrease in performance. This observation reinforces the idea that the UR-GNN's effectiveness stems from its ability to effectively integrate uncertain neighbors and relations information within the entity features. The uncertainty-aware approach of the UR-GNN proves to be more advantageous for the few-shot knowledge graph completion task.
3) By replacing the UR-GNN module with KG embedding methods, a significant and noticeable decrease in performance is observed. This finding demonstrates that GNN-based models are better suited for uncertainty feature representation than KG embedding methods. The UR-GNN leverages the power of GNNs to capture and incorporate uncertainty, providing a more effective solution for few-shot KG completion.
All the observations derived from the table support the effectiveness of the proposed UR-GNN model. Its ability to consider uncertainty, integrate uncertain neighbors and relations, and leverage the power of GNNs proves advantageous in the FKGC task. The experimental results demonstrate that the UR-GNN surpasses previous models and demonstrates superior performance.


To gain a deeper understanding of the impact of UR-GNN layers, we conducted experiments focusing on the number of uncertain relational graph convolution layers. These experiments are depicted in Figure \ref{KGC_layer} (a) and (b), which illustrate the different variations of the UR-GNN.


From the figure, we can observe that:
1) The best results are achieved when the number of uncertain relational graph convolution layers is three. This finding suggests that applying multiple layers can allow the model to access higher-order neighbors, which in turn benefits higher-order message passing. Additionally, the heuristic hyperedges used in the model provide valuable information. The combination of these factors contributes to the superior performance observed with three layers.
2) As the number of UR-GNN layers continues to increase, the performance of the model significantly decreases. This finding indicates that the representation learned by the model becomes indistinguishable when too many layers are employed. The 3-layer UR-GNN effectively learns entity representations with a discernible degree of distinction. However, the 6-layer UR-GNN fails to preserve these distinguishing features.
Based on these observations, it can be deduced that the UR-GNN may encounter the well-known over-smoothing problem when the number of UR-GNN layers becomes too large. Over-smoothing refers to the phenomenon where aggregated node representations become increasingly similar as the number of layers increases, resulting in a loss of discriminative power. Hence, a three-layer UR-GNN configuration appears to be more suitable and effective for the few-shot knowledge graph completion task.
These findings shed light on the optimal configuration of UR-GNN layers, providing guidance for achieving the best performance in few-shot knowledge graph completion experiments. A three-layer UR-GNN strikes the right balance between accessing higher-order information and avoiding over-smoothing, resulting in improved performance and more distinguishable entity representations.

\begin{figure}[t]
  \centering  
 \subfigure[Impacts of the number of random samples.]{
 \includegraphics[width=0.98\linewidth]{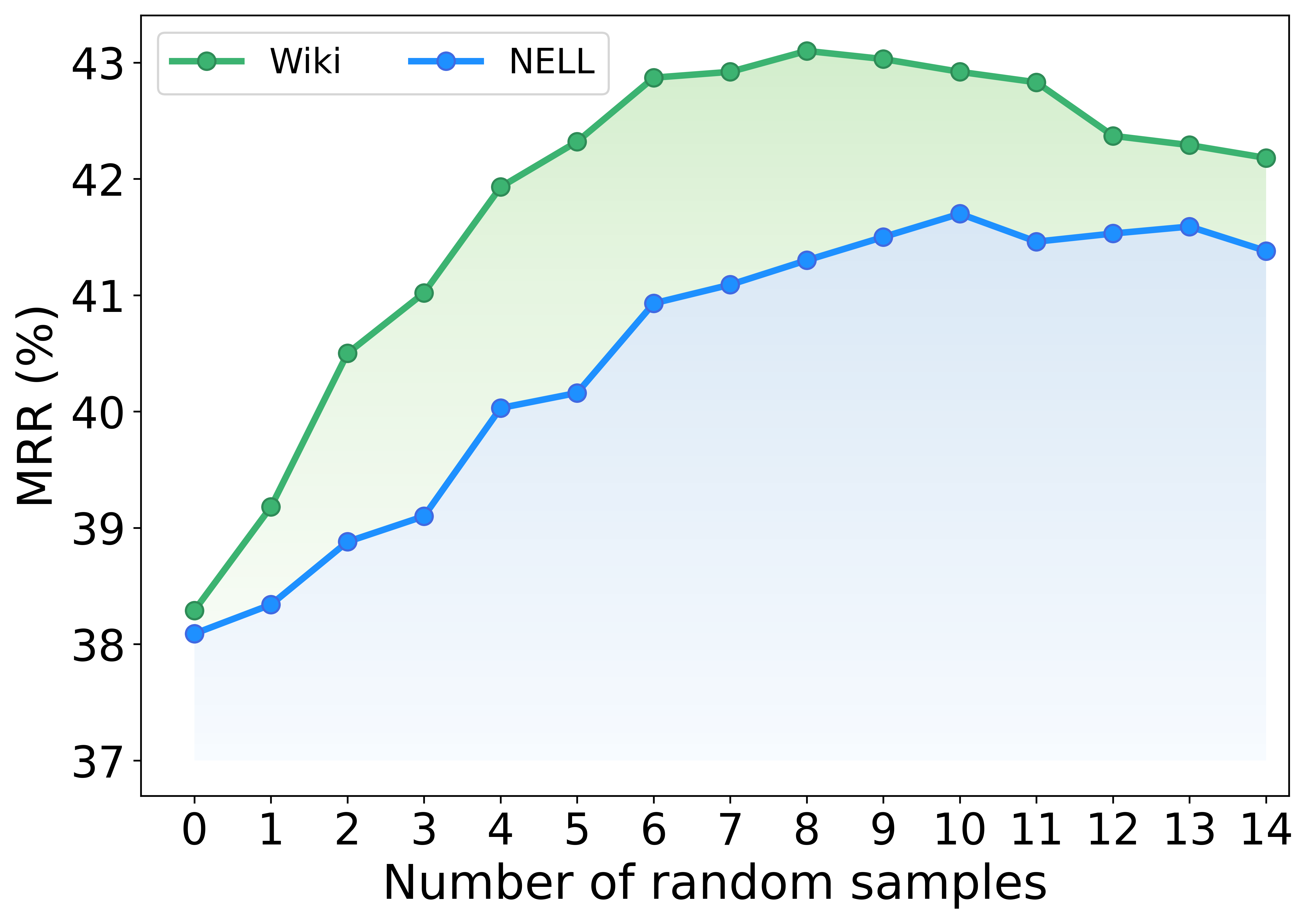}}
    \caption{Impacts of the number of random samples on the NELL and Wiki dataset for the 5-shot situation.}
    \label{samples}
\end{figure}

\subsection{Discussion on Random Samples }
To further investigate the impact of the number of random samples on uncertainty optimization, we conducted experiment with various sample number $m$ settings, shown in Figure \ref{samples}.


By analyzing the results in the figure, we can make the following observations:
1) The plot indicates that as the number of random samples increases, there is an improvement in the model's performance. This finding suggests that incorporating random sampling within the Gaussian distribution helps enhance the model's overall performance. The additional variation provided by random sampling enables the model to capture a broader range of information and make more informed decisions.
2) However, it is interesting to note that beyond a certain threshold, increasing the number of random samples does not lead to further improvements in performance. In fact, in some cases, the performance even slightly decreases. This suggests that there is an optimal number of random samples that strikes a balance between incorporating enough variation and avoiding unnecessary noise. Excessive sampling may introduce noise or irrelevant information, which can interfere with the model's decision-making process and result in a slight decrease in performance.

Based on these observations, we can conclude that for the task of few-shot knowledge graph completion, the most suitable and effective number of random samples is 10 and 8 for the NELL and wiki datasets, respectively. These numbers strike the right balance between incorporating sufficient variation and avoiding excessive noise, resulting in improved performance.
The experimental results strongly support the idea that random sampling within the Gaussian distribution plays a critical role in enhancing the performance of the UR-GNN model for few-shot knowledge graph completion. The identified optimal number of random samples provides guidance to researchers and practitioners working in this domain to achieve better results.

\section{Conclusion}

This paper proposes an uncertainty-aware few-shot knowledge graph completion framework UFKGC, which models the uncertainty of entities and relations, and performs uncertainty optimization. It considers modeling the uncertainty to promote the exploitation of limited samples in few-shot optimization. 
We model the uncertainty of few-shot knowledge graph completion through the uncertainty representation, transferring the origin feature into a Gaussian distribution. 
We also design an uncertainty-aware relational graph neural network (UR-GNN) to better integrate the uncertainty neighbors and relations information for entity features.
We also conduct multiple random sampling within the Gaussian distribution without destroying the original data distribution.
Moreover, to measure the uncertainty for each query instance, we propose uncertainty optimization. 
Without destroying the original data distribution, we can prompt the model to learn consistent knowledge from those generated smooth representations.
The final completion score for each query instance is measured by the designed uncertainty optimization to be robust on the few-shot setting. 
Experiments on two open few-shot knowledge graph completion datasets show that our model beats state-of-the-art techniques with different few-shot sizes.
In future work, we will study how to avoid the influence of unevenly distributed categories of relation on the few-shot knowledge graph completion.

\section*{Acknowledgement}
We thank the anonymous reviewers for their insightful comments and suggestions. 
Jianxin Li is the corresponding author.
The authors of this paper were supported by the NSFC through grant No.U20B2053, 62106059 and the Academic Excellence Foundation of Beihang University for PhD Students.

\bibliographystyle{IEEEtran}
\bibliography{references}
\vspace{-1.2cm}

\begin{IEEEbiography}[{\includegraphics[width=1in,height=1.25in,clip,keepaspectratio]{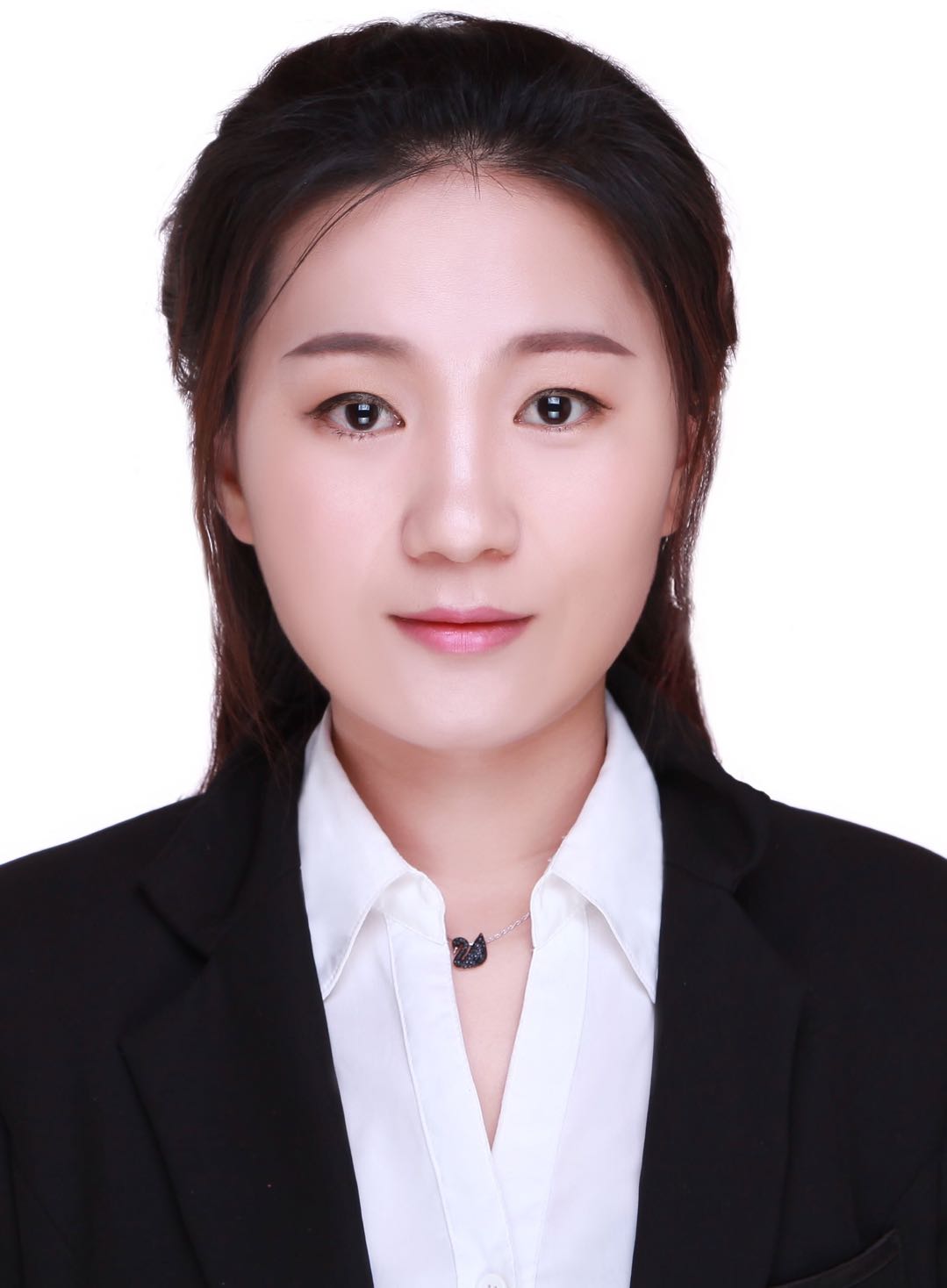}}]{Qian Li} is currently pursuing the Ph.D. degree with the Department of Computer Science and Engineering, Beihang University (BUAA), Beijing, China. Her research interests  include text mining, representation learning, and event extraction.
\end{IEEEbiography}
\vspace{-1.8cm}

\begin{IEEEbiography}[{\includegraphics[width=1in,height=1.25in,clip,keepaspectratio]{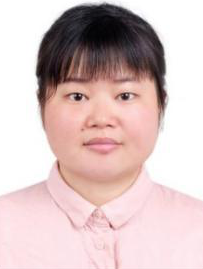}}]{Shu Guo} received Ph.D. degree from the Institute of Information Engineering, Chinese Academy of Sciences. She is currently working at the National Computer Network Emergency Response Technical Team/Coordination Center of China. Her research interests include Knowledge Graph Embedding, Knowledge Acquisition and Web Mining.
\end{IEEEbiography}
\vspace{-1.8cm}

\begin{IEEEbiography}[{\includegraphics[width=1in,height=1.25in,clip,keepaspectratio]{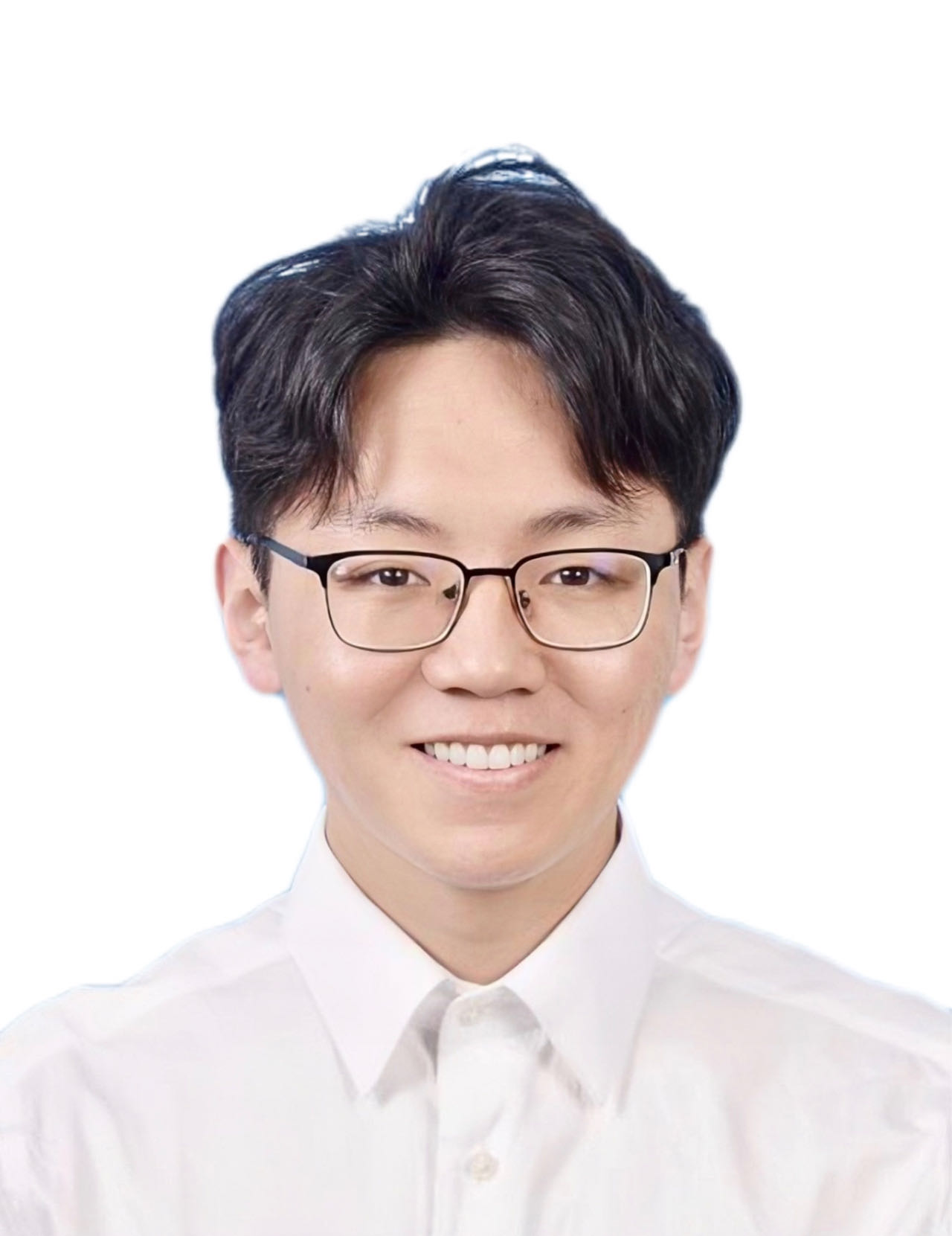}}]{Yinjia Chen} is currently pursuing an MSc. degree at the School of Mathematics, University of Edinburgh. His current research interests include Gaussian Processes, Bayesian Inference, and Knowledge Graph Embedding. 
\end{IEEEbiography}
\vspace{-1.8cm}

\begin{IEEEbiography}[{\includegraphics[width=1in,height=1.25in,clip,keepaspectratio]{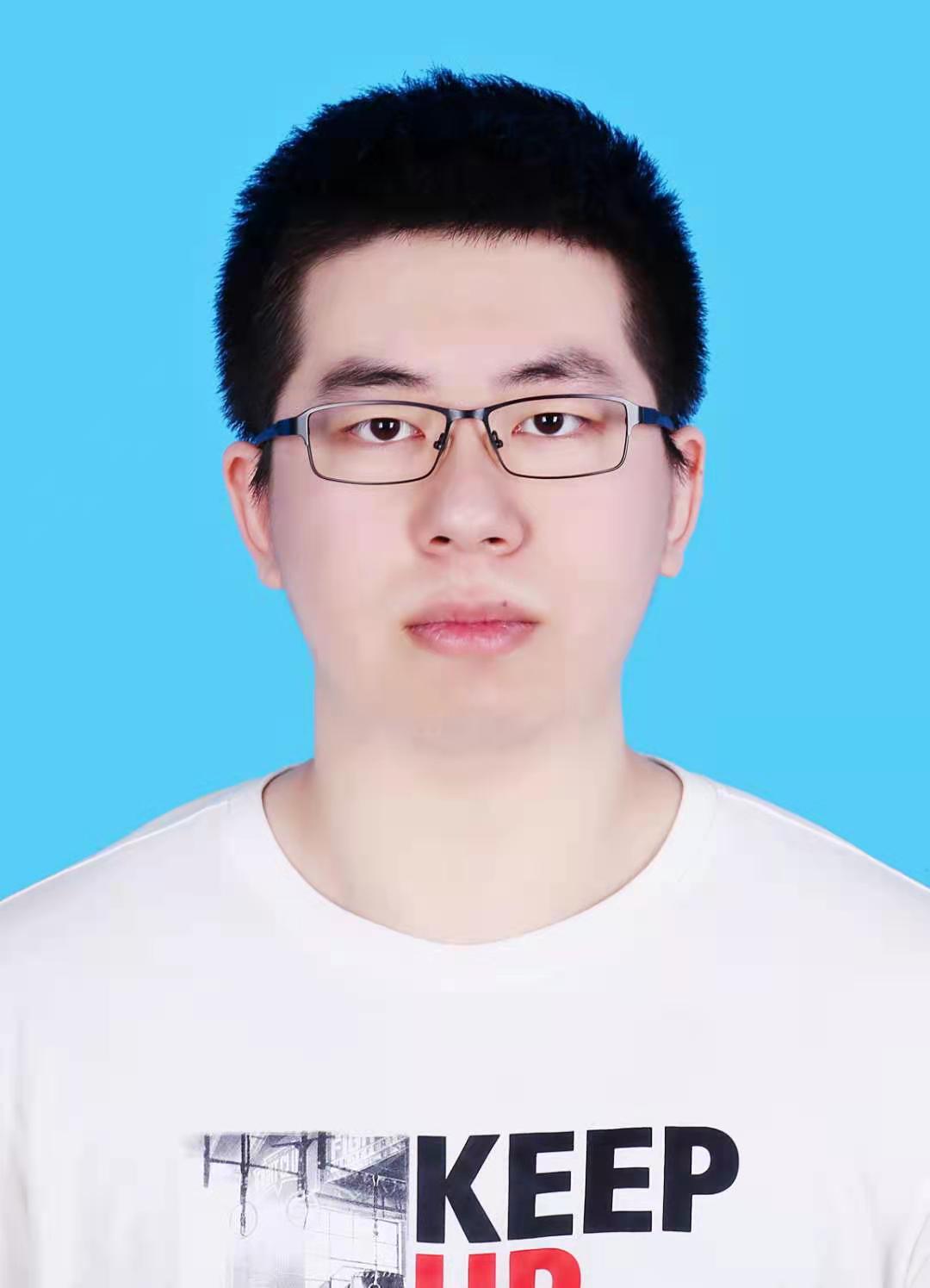}}]{Cheng Ji} is currently a Ph.D. candidate at the Beijing Advanced Innovation Center for Big Data and Brain Computing at Beihang University. His research interests include self-supervised learning and graph representation learning.
\end{IEEEbiography}
\vspace{-1.8cm}

\begin{IEEEbiography}[{\includegraphics[width=1in,height=1.25in,clip,keepaspectratio]{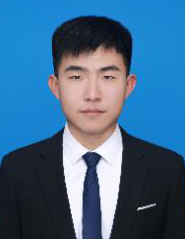}}]{Jiawei Sheng} is currently pursuing the Ph.D. degree in the Institute of Information Engineering, Chinese Academy of Sciences. His current research interests include Information Extraction, Knowledge Graph Embedding and Knowledge Acquisition. 
\end{IEEEbiography}
\vspace{-1.8cm}

\begin{IEEEbiography}[{\includegraphics[width=1in,height=1.25in,clip,keepaspectratio]{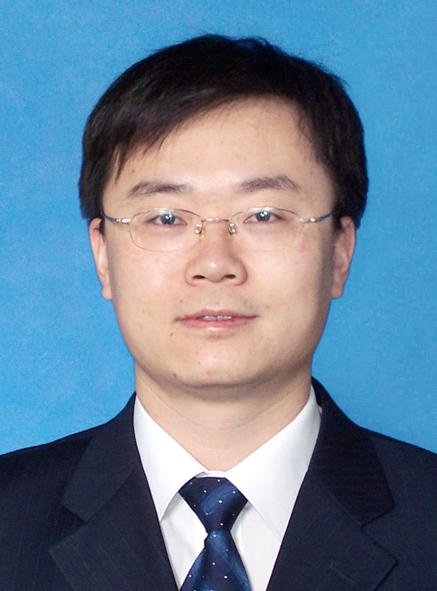}}]{Jianxin Li}
 is currently a Professor with the State Key Laboratory of Software Development Environment, and Beijing Advanced Innovation Center for Big Data and Brain Computing in Beihang University. His current research interests include machine learning, distributed system, trust management and network security. 
\end{IEEEbiography}
\vspace{-1.8cm}

\end{document}